\documentclass[runningheads]{llncs}

\usepackage[T1]{fontenc}
\usepackage{graphicx}
\usepackage{multirow}
\usepackage{amssymb}

\usepackage{CJKutf8}
\usepackage{booktabs}
\usepackage{amsmath}
\usepackage{bbm}
\usepackage{adjustbox}
\usepackage{cite}
\usepackage[most]{tcolorbox}
\usepackage{lipsum} 

\usepackage{marvosym}

\begin{document}
\title{Detection, Classification, and Mitigation of Gender Bias in Large Language Models}

\titlerunning{Detection, Classification, and Mitigation of Gender Bias} 

\author{
Xiaoqing Cheng\inst{1} \and
Hongying Zan\inst{2}\thanks{Corresponding author.} \and
Lulu Kong\inst{1} \and
Jinwang Song\inst{1} \and
Min Peng\inst{3}
}

\authorrunning{X. Cheng et al.} 

\institute{Zhengzhou University, Zhengzhou, China \\
\inst{1}\email{\{xqcheng, kll, jwsong\}@gs.zzu.edu.cn} \\
\inst{2}\email{\{iehyzan\}@zzu.edu.cn} \\
\inst{3}Wuhan University, Wuhan, China, \inst{3}\email{pengm@whu.edu.cn}}
\maketitle    

\begin{abstract}
With the rapid development of large language models (LLMs), they have significantly improved efficiency across a wide range of domains. However, recent studies have revealed that LLMs often exhibit gender bias, leading to serious social implications. Detecting, classifying, and mitigating gender bias in LLMs has therefore become a critical research focus. In the NLPCC 2025 Shared Task 7: Chinese Corpus for Gender Bias Detection, Classification and Mitigation Challenge, we investigate how to enhance the capabilities of LLMs in gender bias detection, classification, and mitigation. We adopt reinforcement learning, chain-of-thoughts (CoT) reasoning, and supervised fine-tuning to handle different Subtasks. Specifically, for Subtasks 1 and 2, we leverage the internal reasoning capabilities of LLMs to guide multi-step thinking in a staged manner, which simplifies complex biased queries and improves response accuracy. For Subtask 3, we employ a reinforcement learning-based approach, annotating a preference dataset using GPT-4. We then apply Direct Preference Optimization (DPO) to mitigate gender bias by introducing a loss function that explicitly favors less biased completions over biased ones. Our approach ranked first across all three subtasks of the NLPCC 2025 Shared Task 7.

\end{abstract}

\section{Introduction}

Large language models, characterized by their extensive parameterization and large-scale training corpora, have been widely applied across various domains, bringing substantial convenience and efficiency~\cite{liu2023g}. However, recent studies have shown that these models often exhibit gender bias due to imbalances in their training data \cite{gallegos2024bias}. Such biases risk perpetuating gender inequality, reinforcing stereotypical social norms, and undermining fairness, thereby leading to serious societal consequences. As a result, detecting, classifying, and mitigating gender bias in large language models has become increasingly important. In response to these challenges, the NLPCC 2025 Task 7 was introduced, focusing on the identification and mitigation of gender bias in Chinese-language data.

Existing bias evaluation methods can be categorized into data-level~\cite{wang2024large}, model-level~\cite{doan2024fairness}, output-level~\cite{banerjee2024all}, and human-involved, each targeting distinct sources and manifestations of bias. These approaches are often constrained by high computational and data annotation costs, and may face challenges in scalability and consistency. Existing methods for gender bias mitigation can be broadly categorized into two types: prompt-based \cite{echterhoff2024cognitive, liu2024zero} approaches and fine-tuning-based~\cite{cheng2024rlrf} approaches. However, these approaches often suffer from limited control over fine-grained bias mitigation and require substantial computational resources, which restrict their scalability and practicality.

Based on the above considerations, we propose different methods to address the three subtasks. For Subtask 1 and Subtask 2, which focus on bias detection and classification\cite{fan2024biasalert, fan2024fairmt}, we employ supervised fine-tuning in combination with chain-of-thought (CoT) reasoning \cite{wei2022chain, kojima2022large}. Specifically, we first inject gender bias-related knowledge into the large language model through supervised fine-tuning. Then, leveraging the model’s reasoning capabilities, we design task-specific prompts that guide the model to decompose complex problems into simpler ones. This staged reasoning process enables the model to think step by step with logical structure, thereby improving the accuracy of its predictions.
For Subtask 3, which involves rewriting sentences to eliminate gender bias \cite{cheng2025biasfilter, li2025fairsteer, chen2025identifying} while preserving the original meaning, we adopt a reinforcement learning \cite{ouyang2022training, lee2023rlaif} approach. We begin by using GPT-4~\cite{rafailov2023direct} to construct a set of preference data pairs from the original training set. Based on this, we train a DPO \cite{rafailov2024r, wang2022exploring} (Direct Preference Optimization) model by designing a loss function \cite{chen2023fast, chen2024learnable} that encourages the model to generate less biased rewritings. To enhance the coverage and diversity of our training data, we further supplement it with samples collected from the CORGI-PM \cite{zhang2023corgipmchinesecorpusgender} dataset.

Experimental results demonstrate the effectiveness of our approach across all three tasks: gender bias detection, classification, and mitigation. Our system achieved No.1 in Subtask 1, Subtask 2, and Subtask 3, as well as the highest overall ranking in NLPCC 2025 Task 7.

Our main contributions are summarized as follows:
\begin{itemize}
    \item We contribute a fine-grained dataset for gender bias detection and classification, as well as a preference dataset constructed via AI feedback. Specifically, in building the preference dataset, we leverage GPT-4 to generate counterexamples. These counterexamples are paired with corresponding unbiased sentences from the training data to form high-quality preference pairs suitable for preference-based optimization.

    \item We introduce a chain-of-thought (CoT) strategy for gender bias detection and classification. By designing task-specific prompts, we guide LLMs to reason step by step, improving both accuracy and interpretability.
    
    \item We propose a novel DPO-based debiasing framework that mitigates gender bias in generated text while preserving semantic intent, enabling LLMs to prefer less biased completions.
\end{itemize}

\section{Related Work}
\subsection{Bias evaluation}

Existing methods for evaluating and classifying gender bias in large language models (LLMs) can be broadly categorized into four categorizes. (1) Data-level approaches primarily focus on identifying and quantifying biases inherent in the training corpora of LLMs. Common strategies include data distribution analysis, data sources analysis, sentiment analysis, and annotation bias analysis. (2) Model-level methods aim to detect biases that arise during the training or inference processes of LLMs, often employing fairness metrics and counterfactual-based evaluations. (3) Output-level evaluations assess how LLMs respond to different demographic groups and whether their outputs maintain fairness and neutrality, typically through techniques such as counterfactual testing, stereotype detection, sentiment and toxicity analysis. (4) Human-involved evaluations incorporate human judgment to capture complex, context-dependent biases in LLM outputs.

\subsection{Bias Mitigation}

Existing gender bias mitigation methods can generally be divided into three major categorizes. (1) prompt-based approaches attempt to reduce bias by crafting prompts that incorporate fairness-related cues, thereby steering the model toward generating more neutral outputs. (2) Fine-tuning-based techniques \cite{zhang2025causal, zhang2023corgipmchinesecorpusgender} focus on retraining models with balanced datasets and utilize a variety of advanced strategies, such as causal debiasing, manipulation of feature subspaces, contrastive learning for self-debiasing, and interventions at the module level. (3) Inference-time debiasing methods concentrate on altering the decoding process during inference, either by constraining token selection at each step or by generating multiple candidate outputs and selecting the least biased among them.

\begin{figure*}[t]
    \centering
    \adjustbox{trim=0.3cm 0.8cm 0.5cm 0.5cm,clip}{
        \includegraphics[width=\linewidth]{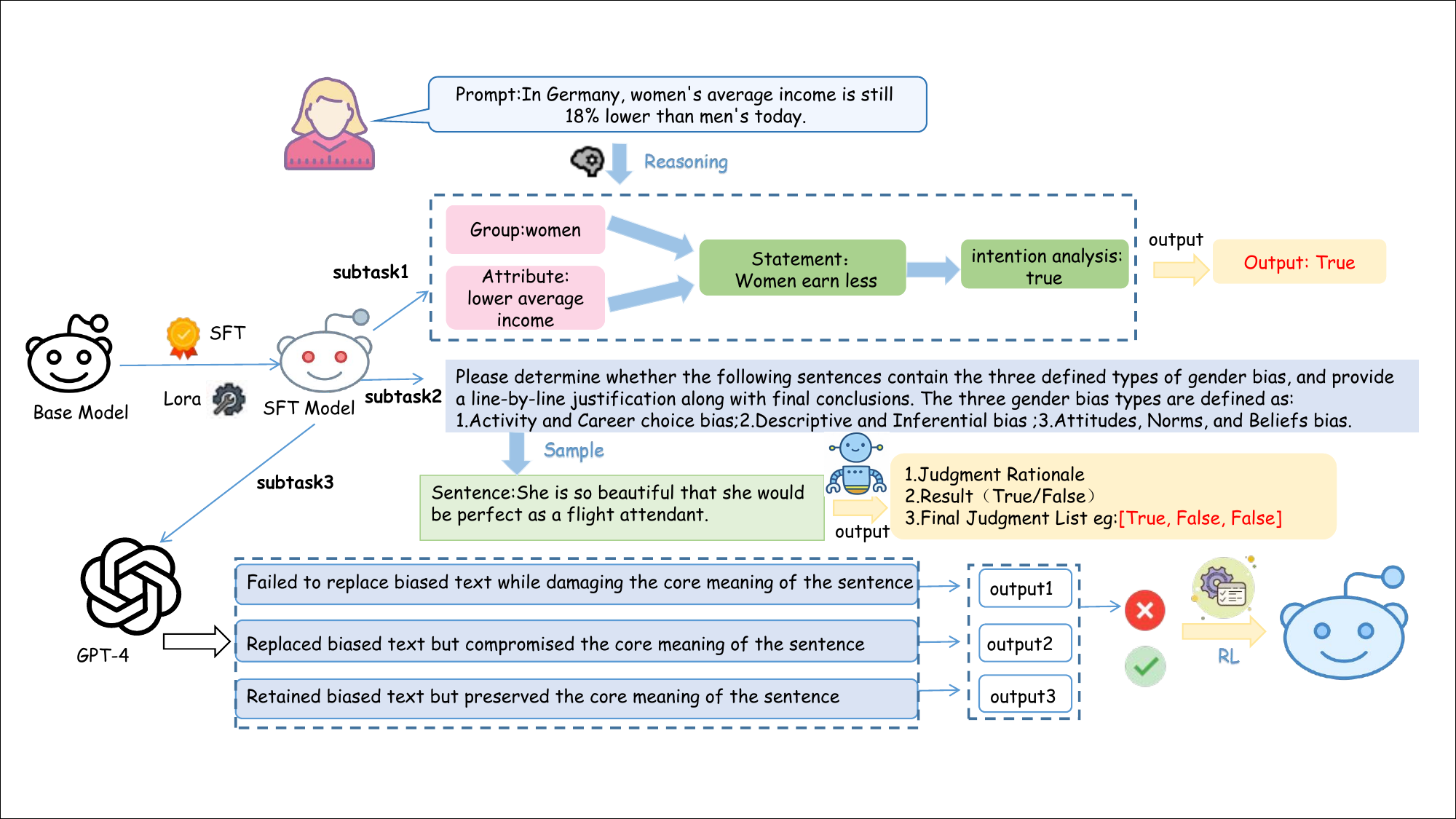}
    }
    \caption{The overall framework for bias detection, classification, and mitigation.}
    \label{fig:2}
\end{figure*}

\section{Methods}

Figure 1 illustrates an overview of our proposed framework for addressing the three subtasks. Supervised fine-tuning serves as the foundation for all three tasks. Building upon this, we design task-specific strategies tailored to the unique characteristics of each subtask. Our approach primarily integrates two distinct techniques: chain-of-thought reasoning, which is applied to Subtask 1 and Subtask 2, and a reinforcement learning-based method, which is mainly employed for Subtask 3. In the following sections, we provide detailed descriptions of each method respectively.

\subsection{Chain-of-Thought Reasoning}
\paragraph{Task1: Bias Detection.}
For Subtask 1, we randomly sample a portion of the CORGI-PM dataset \cite{zhang2023corgipmchinesecorpusgender} and combine it with the official training data provided by the organizers to construct the raw dataset for fine-tuning the large language model. Following~\cite{fan2025biasguard}, during training and inference, we instruct the model to respond in a structured format comprising three steps:

Groups and Attribute Identification: The model is first required to identify the social group mentioned in the sentence (e.g., “women” in the example) and the associated attribute assigned to the group (e.g., “lower average income”).

Bias Judgment: Next, the model is asked to determine whether describing the identified group with the given attribute (e.g., “women earn less”) constitutes a biased statement (e.g., answer: “true”).

Agreement Analysis and Label Assignment: Finally, the model must assess whether the sentence expresses agreement with the biased statement identified in step 2. If the sentence aligns with the biased implication, the output label is set to “True”; otherwise, if the sentence opposes or questions the implication, the label is set to “False”.

\paragraph{Task2: Bias Classification.}
For Subtask 2, which involves a multi-label classification task, it is challenging for the model to simultaneously make accurate predictions across multiple categories, especially when it may not fully understand the semantic distinctions among the labels. To address this, our prompt is carefully designed to enhance label comprehension and step-by-step reasoning.

Specifically, we first define the meaning of the three labels involved in the task. Then, we ask the model to independently assess whether the given sentence belongs to each label. For each assessment, the model is required to provide a justification before producing a binary decision (i.e., whether the label applies). In the final step, the model is instructed to synthesize the outcomes of the three individual judgments and generate a combined overall answer.

\subsection{Reinforcement Learning-based Method}

For Subtask 3, we adopt a reinforcement learning \cite{chen2025diffpo, chen2024pad, zhang2025persona} approach by training a Direct Preference Optimization (DPO) policy model on a custom-constructed dataset. The goal is to rewrite sentences in a way that maximally reduces gender bias while preserving the original semantic intent.

\paragraph{Framework.}
Our approach trains the large language model (LLM) using a preference-based loss function that encourages the generation of responses aligned with human values. The loss maximizes the log-likelihood of tokens in completions that are considered less gender-biased, more respectful, and more consistent with human preferences, while minimizing the likelihood of completions that exhibit gender bias, disrespect, or harmful content.
The debiasing loss function is defined as follows:
\begin{equation}
\begin{split}
\mathcal{L}_{\text{DPO}}(\pi; \pi_{\text{ref}}) 
= -\mathbb{E}_{(x, y_w, y_l) \sim D} \Big[
\log \sigma \Big(
\beta \log \frac{\pi(y_w \mid x)}{\pi_{\text{ref}}(y_w \mid x)}
- \beta \log \frac{\pi(y_l \mid x)}{\pi_{\text{ref}}(y_l \mid x)}
\Big) \Big]
\end{split}
\end{equation}

where $\pi$ represents the target policy, and $\pi_{\text{ref}}$ denotes the reference policy. Each training instance $(x, y_w, y_l)$ is sampled from the Fairness Preference Dataset $D$, where $x$ is the prompt, $y_w$ and $y_l$ are the preferred and less preferred responses, respectively. $\beta$ controls the degree of divergence of $\pi$ from the reference policy $\pi_{\text{ref}}$. To prevent the model from diverging excessively from the original data distribution, a reference model is incorporated through a Kullback–Leibler (KL) divergence term. This training objective penalizes the model when it produces biased outputs and rewards it for generating less biased alternatives, thereby encouraging the production of respectful and non-harmful language.

\paragraph{Dataset.}
To better guide the DPO model in learning human-aligned preferences, we carefully constructed a dedicated dataset for DPO training. Specifically, for each prompt in the training set, we leverage GPT-4 to generate sentences following predefined rules. The generated sentence was treated as the dispreferred response, while the human-edited version from the training data served as the preferred response. These preference pairs were then used to train the DPO model.

The dataset construction process is illustrated in Figure~\ref{fig:2}. We adopted GPT-4 to generate counterfactual examples by replacing parts of the original sentence, aiming to produce contrastive pairs that align with the goals of our task. In particular, we consider three types of counterfactuals: (i) biased content is not replaced and the core meaning of the sentence is distorted; (ii) biased content is replaced, but the core meaning is still distorted; and (iii) biased content is retained, but the core meaning is preserved. To ensure coverage of these cases, we designed different prompting strategies to guide GPT-4 in generating high-quality counterfactual responses. This design allows the DPO model to better distinguish between factual and counterfactual examples, thereby improving its ability to rewrite biased sentences in a way that reduces gender bias while maintaining semantic fidelity.

\section{Experiments}
\subsection{Datasets and Metrics.} 
\paragraph{Datasets.} We conduct our experiments using the dataset provided in Shared Task 7 of NLPCC 2025. In addition, we randomly sample a subset of data from CORGI-PM to supplement our training set. The statistics of the three tasks are summarized in Table~\ref{tab:task_stats}. For each task, the data is divided into training, validation, and test sets.

\begin{table}[ht]
\renewcommand{\arraystretch}{1.5} 
\setlength{\tabcolsep}{12pt} 
\caption{Dataset statistics for each subtask.}
\centering
\resizebox{\textwidth}{!}{ 
\begin{tabular}{lrrrr}
\hline
\textbf{Task} & \textbf{Train} & \textbf{Valid} & \textbf{Test} & \textbf{Total} \\
\hline
SubTask1 & 12224 & 1032 & 200 & 12924 \\
SubTask2 & 4872  & 516  & 200 & 5472  \\
SubTask3 & 3672  & 516  & 200 & 4372  \\
\hline
\end{tabular}
}

\label{tab:task_stats}
\end{table}

\paragraph{Evaluation Metrics.} 

Task 1: Bias Detection.
This task is formulated as a binary classification problem. Given a sentence, the goal is to determine whether it contains gender bias. 

We adopt a commonly used evaluation metric: F1-score, which are defined as follows:

\begin{equation}
\text{F1-score} = \frac{2 \cdot \text{Precision} \cdot \text{Recall}}{\text{Precision} + \text{Recall}}
\end{equation}

\noindent
where Precision denotes the fraction of correctly predicted positive instances among all predicted positives, and Recall denotes the fraction of correctly predicted positive instances among all actual positives.

Task 2: Gender Bias Classification.
This task focuses on categorizing identified gender biases into three types:  Activity and Career Choices (AC), Gender Stereotyped Descriptions and Inductions (DI), Expressed Gender-stereotyped Attitudes, Norms and Beliefs (ANB)

We adopt both class-wise and macro-averaged evaluation metrics: F1-score. Class-wise metrics for class $i$ are defined as:

\begin{equation}
\text{F1}_i = \frac{2 \cdot \text{Precision}_i \cdot \text{Recall}_i}{\text{Precision}_i + \text{Recall}_i}
\end{equation}

\noindent
where \( i \in \{1, 2, 3\} \) indexes the three predefined gender bias types.

The macro-averaged scores are computed as the unweighted mean over all $K$ classes:

\begin{equation}
\text{Macro-F1} = \frac{1}{K} \sum_{i=1}^{K} \text{F1}_i
\end{equation}
In this task, \( K = 3 \), corresponding to the three gender bias categories.

Task 3: Gender Bias Mitigation.
This task focuses on rewriting biased sentences to mitigate gender bias while preserving the core meaning of the original text. The goal is to generate unbiased variants of the input sentences.

We adopt a commonly used evaluation metric: BLEU.

BLEU evaluates the overlap of n-grams between the generated sentence and one or more reference sentences. It emphasizes precision and includes a brevity penalty to penalize overly short outputs.

\subsection{Baselines.} 
\paragraph{(1) LLM Zero-shot.} 
We evaluate several large language models in a zero-shot setting on all three tasks to serve as baselines. The models include Chinese Tiny LLM (2B)~\cite{du2024chinese}, Qwen2.5-Instruct (7B)~\cite{hui2024qwen2}, Yi-1.5 (6B)~\cite{arias2020yi}, and GPT-4o.

\paragraph{(2) SFT Training.}
We fine-tune the base models on the training datasets provided for the three tasks. The performance is compared with our proposed chain-of-thought prompting method and reinforcement learning-based method.

\paragraph{(3) Reward-guided Generation.}
We adopt the reward-guided decoding framework. Specifically, the loss for reward model training is defined as:
\begin{equation}
\mathcal{L}_{\text{RM}}(x, y_w, y_l; \theta) = \log \sigma\left(r([x, y_w]) - r([x, y_l])\right)
\end{equation}

where $\theta$ is the parameter set of the reward model, $\sigma(\cdot)$ is the sigmoid function, and $r([x, y])$ denotes the scalar reward for a given input-response pair. 

Given the context $x_{<t}$ and timestep $t$, the reward-guided token scoring function is defined as:
\begin{equation}
s(v, x_{<t}) = \text{LM}(v \mid x_{<t}) + w \cdot r([x_{<t}, v])
\end{equation}

where $\text{LM}(v \mid x_{<t})$ is the model's likelihood for token $v$, and $w$ is a scalar weight for the reward.

\paragraph{(4) BiasDPO.}
We follow the setup proposed in BiasDPO \cite{allam2024biasdpo}, where an additional debiasing dataset is constructed and used to train a DPO model. The resulting DPO model is then used on all three tasks.

\paragraph{(5) Without Data Expansion.}
To assess the effect of external data, we fine-tune the base model using only the official NLPCC 2025 task dataset, without incorporating any supplementary samples from the CORGI-PM dataset.

\subsection{Experimental Settings.}
All experiments are conducted using Qwen2.5-7B-Instruct as the base model. We used the LLaMA Factory~\cite{li2025drafts} framework for training. The training configurations for SFT and DPO are shown in Table~\ref{tab:hyperparams}.

For SFT Training and Reward-guided Generation baselines, we also adopt Qwen2.5-7B-Instruct as the base model. In Reward-guided Generation, where a reward model is required, we use Meta-LLaMA3-8B-Instruct~\cite{alsariera2020ai} as the backbone to train the reward model. The training configuration for the reward model is shown in Table~\ref{tab:hyperparams}.

\begin{table}[h]
\centering
\caption{Hyperparameter settings: left for SFT training, right for DPO training and RM training.}

\begin{minipage}{0.48\linewidth}
\centering

\begin{tabular}{ll}
\toprule
\textbf{Hyper-parameter} & \textbf{Value} \\
\midrule
Lora Alpha         & 16 \\
Lora Rank          & 8 \\
Optimizer          & AdamW \\
Train Batch Size   & 1 \\
Train Epochs       & 2 \\
Learning Rate      & $1 \times 10^{-5}$ \\
Max Gradient Norm  & 0.3 \\
Warmup Ratio       & 0.03 \\
Max Sequence Length& 1024 \\
\bottomrule
\end{tabular}
\end{minipage}
\hfill
\begin{minipage}{0.48\linewidth}
\centering

\begin{tabular}{ll}
\toprule
\textbf{Hyper-parameter} & \textbf{Value} \\
\midrule
Lora Alpha         & 32 \\
Lora Rank          & 16 \\
Optimizer          & AdamW \\
Train Batch Size   & 1 \\
Train Epochs       & 2 \\
Learning Rate      & $8 \times 10^{-6}$ \\
Max Gradient Norm  & 0.3 \\
Warmup Ratio       & 0.03 \\
Max Sequence Length& 1024 \\
\bottomrule
\end{tabular}
\end{minipage}
\label{tab:hyperparams}
\end{table}

\subsection{Experimental Results and Analysis.}
In this section, we present a comprehensive comparison between our proposed method and various baselines, which constitutes the core results of our main experiments. In addition, we conduct a series of ablation studies to validate the effectiveness of each component in our framework. Finally, we perform an error analysis by examining typical failure cases, aiming to gain deeper insights into the challenges of the task and to inform future improvements.

\begin{table}[ht]
\centering
\renewcommand{\arraystretch}{1.2}
\setlength{\tabcolsep}{10pt}
\caption{Main results on the validation sets of the three subtasks.}
\begin{adjustbox}{width=0.95\linewidth}
\begin{tabular}{llccc}
\toprule
\textbf{Methods} & \textbf{Models} & \textbf{Subtask1} & \textbf{Subtask2} & \textbf{Subtask3} \\
\midrule
\multirow{4}{*}{LLM Zero-shot} 
    & Chinese Tiny LLM (2B)      & 0.24 & 0.32 & 0.154 \\
    & Qwen2.5-Instruct (7B)      & 0.69 & 0.58 & 0.218 \\
    & Yi-1.5 (9B)                & 0.68 & 0.58 & 0.215 \\
    & GPT-4o                     & 0.82 & 0.67 & 0.259 \\
\midrule
\multirow{1}{*}{SFT Training}      
    & Qwen2.5-Instruct (7B)      & 0.72 & 0.59 & 0.248 \\

\multirow{1}{*}{ARGS}             
    & Qwen2.5-Instruct (7B)      & 0.83 & 0.61 & 0.269 \\

\multirow{1}{*}{BiasDPO}         
    & Qwen2.5-Instruct (7B)      & 0.75 & 0.63 & 0.223 \\

\multirow{1}{*}{Ours}            
    & Qwen2.5-Instruct (7B)      & \textbf{0.87} & \textbf{0.68} & \textbf{0.286} \\
\midrule
w/o Dataset Expansion  & Qwen2.5-Instruct (7B)      & 0.85 & 0.65 & 0.265 \\
w/o CoT                & Qwen2.5-Instruct (7B)      & 0.72 & 0.59 & - \\
w/o RL                 & Qwen2.5-Instruct (7B)      & - & - & 0.248 \\
\bottomrule
\end{tabular}
\end{adjustbox}

\label{tab:main_results}
\end{table}

\paragraph{Main Experimental Results.}
We compare all baseline methods with our proposed approach on the validation sets, and report the results in Table~\ref{tab:main_results}. Compared with several strong baselines, our method achieves the best performance across all three tasks, demonstrating its effectiveness. Interestingly, we observe that the ARGs-based method also performs competitively, suggesting that reinforcement learning-based strategies are generally effective for bias mitigation, and that the benefit is not limited to DPO-style optimization. However, we also find that BiasDPO, despite being trained with a reinforcement learning objective, does not perform as well. Upon closer examination, we attribute this to the dataset used in BiasDPO, which may not be well aligned with the specific settings of our tasks. This finding indirectly supports the effectiveness of our approach to constructing contrastive pairs via GPT-4 sampling, which ensures task relevance and better supervision.

\paragraph{Ablation Study.}
To investigate the contribution of different components in our method, we conduct ablation studies by individually removing the following modules: (1) the data expansion component, (2) the chain-of-thought (CoT) reasoning mechanism, and (3) the reinforcement learning (RL) optimization. The results are presented in the lower part of Table~\ref{tab:main_results}. We observe that removing the data expansion module leads to an average performance drop across all three tasks. Furthermore, when the CoT module and the reinforcement learning method are removed, the performance on the corresponding tasks also decreases. These results highlight the effectiveness of both the CoT reasoning and reinforcement learning components in our proposed framework.

\paragraph{More Results.}
We further compared the performance of our proposed method combined with different base models, including Qwen2.5-3B and Baichuan2-13B \cite{yang2023baichuan}. The experimental results are presented in Table~\ref{tab:results}.

\begin{table}[htbp]
\centering
\caption{More results of different models on three subtasks.}
\label{tab:results}
\begin{tabular}{llccc}
\hline
\textbf{Methods} & \textbf{Models} & \textbf{Subtask1} & \textbf{Subtask2} & \textbf{Subtask3} \\
\hline
\multirow{2}{*}{LLM Zero-shot} & Baichuan2-13B & 0.58 & 0.54 & 0.198 \\
                                & Qwen2.5-3B    & 0.61 & 0.52 & 0.193 \\
\hline
\multirow{2}{*}{Ours}           & Baichuan2-13B & 0.69 & 0.66 & 0.232 \\
                                & Qwen2.5-3B    & 0.65 & 0.65 & 0.227 \\
\hline
\end{tabular}
\end{table}

\paragraph{Error Analysis.}
Table~\ref{tab:error_types} presents the error distribution on Subtask 2. We randomly sampled 50 error cases and manually analyzed each one to identify its error type. Our analysis revealed that, Among the three categories, the error rate for gender-stereotypical activities and career choices (AC) is 26\%, for gender-stereotypical descriptions and inferences (DI) is 38\%, and for gender-stereotypical attitudes, norms, and beliefs (ANB) is 36\%. These results suggest that instances involving descriptive and inferential gender stereotypes (DI) pose greater challenges for computational models, potentially due to their implicit and context-dependent nature.

\begin{table}[htbp]
\centering
\caption{Error Analysis.}
\label{tab:error_types}
\begin{tabular}{lc}
\hline
\textbf{Error Type} & \textbf{Ratio} \\
\hline
Activity and Career Choices (AC)  & 26\% \\
Gender Stereotyped Descriptions and Inductions (DI)  & 38\% \\
Expressed Gender-stereotyped Attitudes, Norms and Beliefs (ANB)  & 36\% \\
\hline
\end{tabular}
\end{table}

\section{Conclusion}

In this paper, we present a framework to address gender bias in large language models (LLMs), targeting bias detection, classification, and mitigation tasks in NLPCC 2025 Shared Task 7. By integrating chain-of-thought (CoT) reasoning, supervised fine-tuning, and reinforcement learning via Direct Preference Optimization (DPO), our method effectively enhances the reasoning and debiasing capabilities of LLMs. Notably, we construct a high-quality preference dataset and utilize GPT-4 to guide bias-aware learning. Experimental results demonstrate that our method achieves state-of-the-art performance across all three subtasks, ranking first in the competition. These findings highlight the potential of combining structured reasoning and preference-based optimization to mitigate social biases in LLMs, paving the way for more fair and responsible language generation systems.

 \bibliographystyle{splncs04}
 \bibliography{1}

\end{document}